\documentclass[conference]{IEEEtran}
\IEEEoverridecommandlockouts
\usepackage{cite}
\usepackage{amsmath,amssymb,amsfonts}
\usepackage{algorithmic}
\usepackage{graphicx}
\usepackage{textcomp}
\usepackage{xcolor}
\usepackage{url}
\usepackage{hyperref}
\hypersetup{
    colorlinks=true,
    linkcolor=blue,
    filecolor=magenta,      
    urlcolor=cyan,
}
\def\BibTeX{{\rm B\kern-.05em{\sc i\kern-.025em b}\kern-.08em
    T\kern-.1667em\lower.7ex\hbox{E}\kern-.125emX}}
\begin{document}
\title{Direct Servo Control from In-Sensor CNN Inference \\with A Pixel Processor Array
\thanks{This work is accepted by ICRA2021 workshop On and Near-sensor Vision Processing, from Photons to Applications (ONSVP). This work was supported by UK EPSRC EP/M019454/1, EP/M019284/1, EPSRC Centre for Doctoral Training in Future Autonomous and Robotic Systems: FARSCOPE and China Scholarship Council (No. 201700260083). }
}

\makeatletter
\newcommand{\linebreakand}{%
  \end{@IEEEauthorhalign}
  \hfill\mbox{}\par
  \mbox{}\hfill\begin{@IEEEauthorhalign}
}
\makeatother

\author{
\IEEEauthorblockN{Yanan Liu}
\IEEEauthorblockA{\textit{Bristol Robotics Laboratory} \\
\textit{University of Bristol}\\
Bristol, UK \\
yanan.liu@bristol.ac.uk}
\and
\IEEEauthorblockN{Jianing Chen}
\IEEEauthorblockA{\textit{School of Electrical \& Electronic Engineering} \\
\textit{University of Manchester}\\
Manchester, UK \\
jianing.chen@manchester.ac.uk}
\and
\IEEEauthorblockN{Laurie Bose}
\IEEEauthorblockA{\textit{Visual Information Labratory} \\
\textit{University of Bristol}\\
Bristol, UK \\
lauriebose@gmail.com}
\linebreakand
% \and
\IEEEauthorblockN{Piotr Dudek}
\IEEEauthorblockA{\textit{School of Electrical \& Electronic Engineering} \\
\textit{University of Manchester}\\
Manchester, UK \\
p.dudek@manchester.ac.uk}
\and
% \linebreakand
\IEEEauthorblockN{Walterio Mayol-Cuevas}
\IEEEauthorblockA{\textit{Visual Information Laboratory} \\
\textit{University of Bristol}\\
Bristol, UK \\
walterio.mayol-cuevas@bristol.ac.uk}
}

\maketitle

\begin{abstract}

This work demonstrates direct visual sensory-motor control using high-speed CNN inference via a SCAMP-5 Pixel Processor Array (PPA). We demonstrate how PPAs are able to efficiently bridge the gap between perception and action. A binary Convolutional Neural Network (CNN) is used for a classic rock, paper, scissors classification problem at over 8000 FPS. Control instructions are directly sent to a servo motor from the PPA according to the CNN's classification result without any other intermediate hardware. 
\end{abstract}


\section{Introduction}

Real-time image capture, processing, and decision making with low-power consumption are essential for next-generation smart sensors. If a complete  neural network inference can be carried out on-sensor, it allows the sensor output to be reduced from entire images to only a small amount of meaningful extracted information used to determine actions. This  considerably reduces the communication and energy costs with auxiliary devices. In-sensor visual computing is an area of growing interest \cite{zhou2020near}. The SCAMP vision system is such an in-sensor visual computing device that performs image processing while sensing and without recording images and while using low power consumption \cite{chen2018scamp5d,carey2013100}. This is in contrast to a conventional vision system and other non on-sensor computing cameras, where the understanding of the visual inputs happens only after the visual data is transmitted to, and processed by, CPUs and or GPUs, resulting in extra latency and power consumption. This paper illustrates how the SCAMP vision system can operate as an  "edge AI" sensor, enabling on-sensor CNN inference and direct action execution with a servo motor, as shown in Fig. \ref{fig:scamp-servo}.

\begin{figure}[t]
\centering
\includegraphics[width=2.5in]{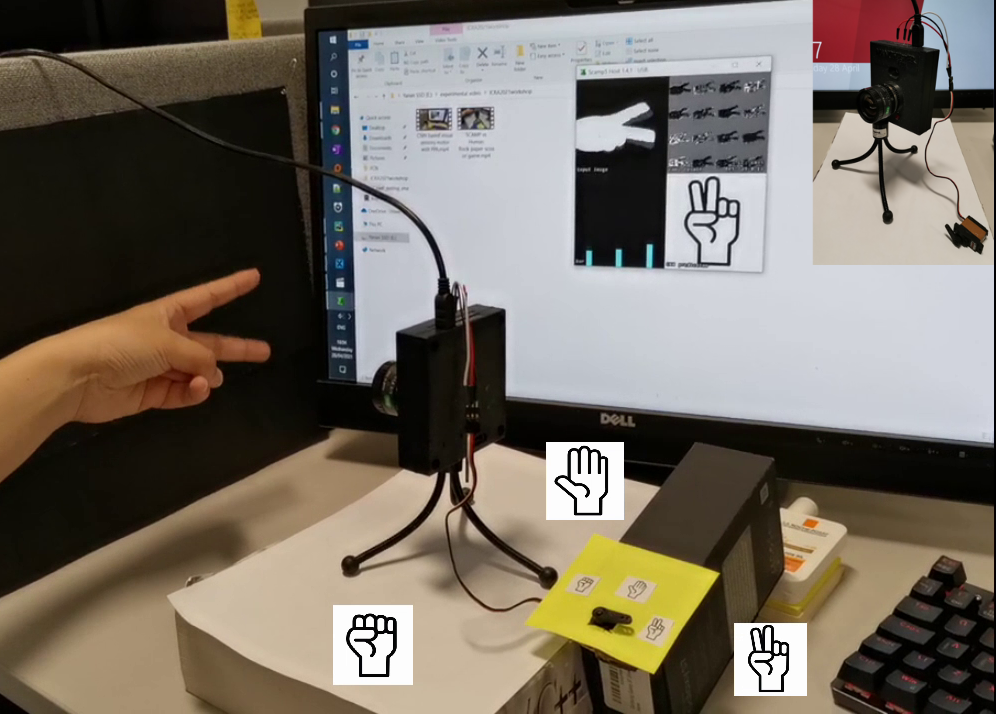}
\caption{A servo motor is directly driven by the SCAMP vision system without additional intermediate hardware. With the machine vision computing results from the SCAMP vision system, the servo can be instructed accordingly. An Align DS425M digital servo is used in this work. 
 }
\label{fig:scamp-servo}
\end{figure}

\begin{figure}[t]
\centering
\includegraphics[width=2.5in]{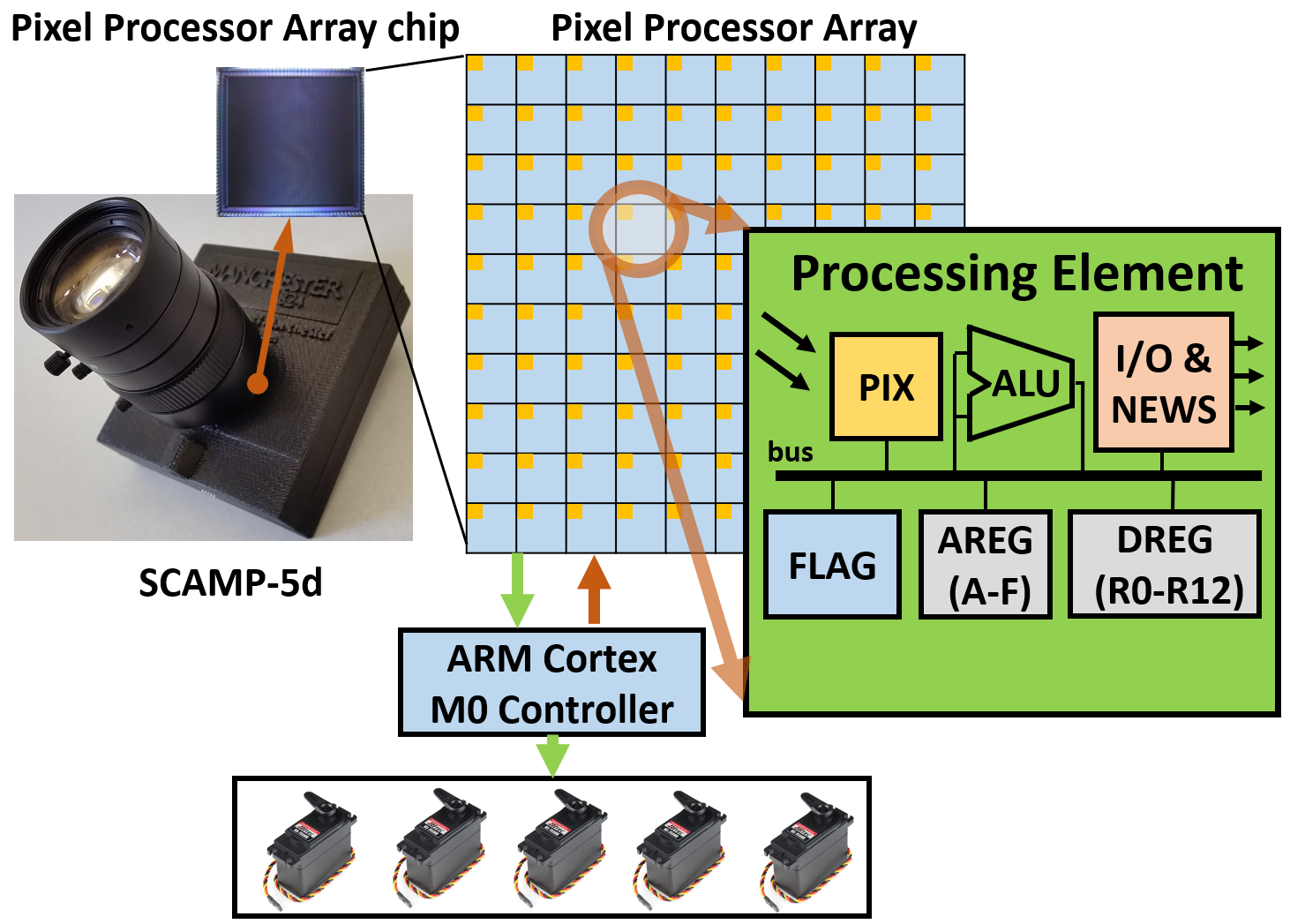}
\caption{The CNN inference in this work is wholly performed on the PPA and the servos can be controlled through GPIO from the ARM M0 controller. A maximum of 5 servos is supported to set up a more comprehensive sensory-motor system.}
\label{fig:scamp-schematic_map}
\end{figure}

\begin{figure*}[t]
\centering
\includegraphics[width=6.8in]{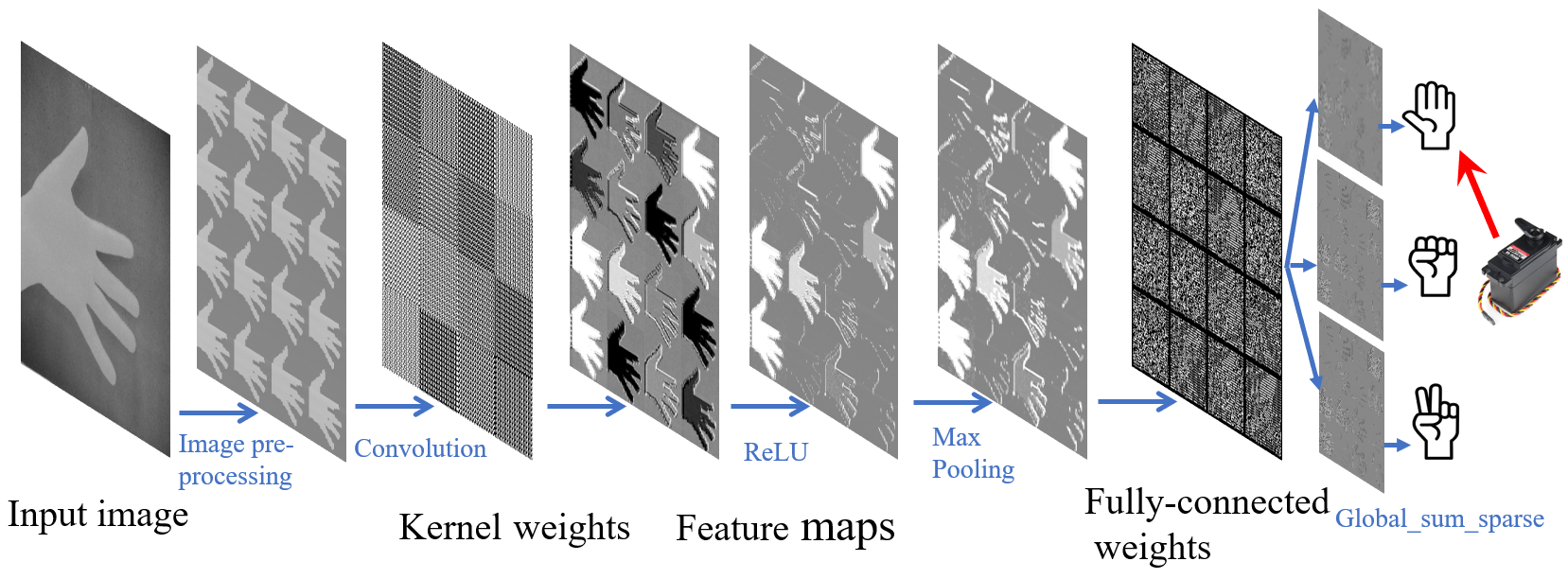}
\caption{Binary CNN inference on the PPA for hand gesture recognition with direct servo control. Image convolution, ReLU activation function, max pooling and fully connected are performed in parallel on the sensing plane.}
\label{fig:scamp-inference}
\end{figure*}

\begin{figure*}[h]
\centering
\includegraphics[width=5.8in]{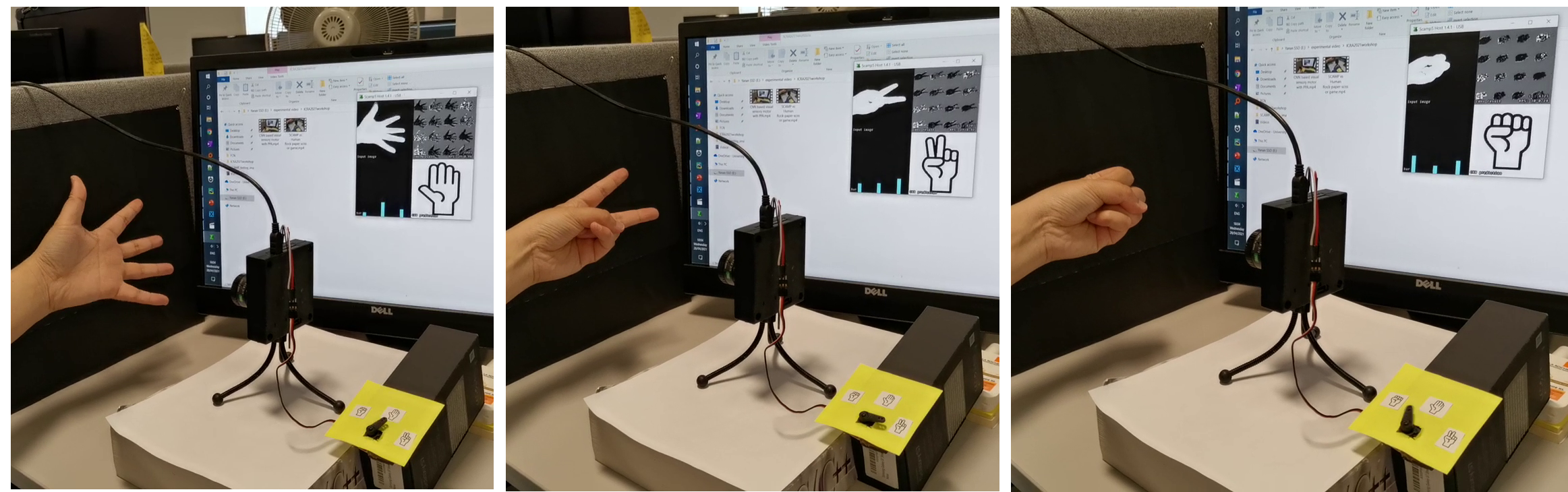}
\caption{CNN inference for hand gesture recognition with a servo as the result indicator.}
\label{fig:scamp-hand}
\end{figure*}

\section{Binary CNN on the PPA}

During recent years, there have been many types of CNN binarization methods proposed \cite{simons2019review} to minimise CNN model size, improve inference efficiency, and reduce power cost. However, embedded hardware usually has its own dedicated hardware design which is not universally applicable to all forms of binary neural network. To deal with this problem for PPA, this work trains and deploys a binary CNN (with kernel weights and fully connected weights of {-1, 1}) onto PPA to fully take advantage of its in-sensor parallel computing feature. Binary weights for both convolutional and fully connected layers are stored in the digital registers (DREG) of the SCAMP processor array (Fig. \ref{fig:scamp-schematic_map}). The processing scheme is illustrated in detail in Fig. \ref{fig:scamp-inference}. Firstly, the analogue PIX registers acquire light signals, forming the image to be passed through the network. Then the image is binarised,  resized and replicated into a dimension of $16\times64\times64$ using on-sensor processing. Different convolution operations are then carried out simultaneously on these 16 replicated images. Then the activation function ReLU sets negative values to zero. Followed by the $2\times2$ maxpooling, a maximum value is selected and replicated within each $2\times2$ block. Lastly, the fully connected layer is implemented by activating each label's AREG and then using the SCAMP built-in global summation function to obtain the approximated value for each label. The index for the biggest summation represents the final CNN inference result \cite{liu2020bmvc}.

% Without bothering to perform analogue-digital conversion as conventional sensors do, the analogue signal can be directly processed within AREG and DREG.

\section{Experiments}

Fig. \ref{fig:scamp-servo} and \ref{fig:scamp-hand} shows the test-bed for the sensory-motor system. The USB link from the computer is used for power supply and transmission of visualisation data. We extended our previous work \cite{liu2020bmvc,bose2020} by adding the implementation of the direct servo control using the binary CNN inference results. This work is based on the SCAMP-collected hand gesture dataset which consist of gestures for 'rock', 'paper', and 'scissors'. The servo motor is controlled through pulse-width modulation on the ARM Cortex M0 processor embedded in the SCAMP system, where the computing results are shared from the PPA. We perform the experiments on three types of hand gesture recognition. This basic setup is used here as a test-bed for bridging perception to action. After the implementation of CNN on the SCAMP, the experimental results (details can be seen from the video results) offer recognition accuracy of $>$97\% with a latency of 121$\mu$s and max theoretical throughput of 8264 FPS \cite{liu2020bmvc}. Note the data transmission to a computer for visualisation slows down the overall FPS, which can be noticed from the experimental video. In addition, the reaction of a servo to inference results is slower than the visualisation updates which results from the low control frequency of only 333 Hz. The experimental video can be seen at  \textit{\url{https://youtu.be/gHcuv275Qrk}} with a $\times20$ slow motion available from \textit{\url{https://youtu.be/SAMsIqqCZ7I}}

\section{Conclusions and future work}

A visual sensory-motor scheme based on the PPA is presented in this work, where a servo is directly controlled by the focal-plane CNN inference without additional computing units. We believe that closing the gap between perception and action, through in-sensor computing strategies, is essential for agile and efficient robotic systems.

\bibliographystyle{IEEEtran}
\bibliography{ref}

\begin{thebibliography}{1}
\providecommand{\url}[1]{#1}
\csname url@samestyle\endcsname
\providecommand{\newblock}{\relax}
\providecommand{\bibinfo}[2]{#2}
\providecommand{\BIBentrySTDinterwordspacing}{\spaceskip=0pt\relax}
\providecommand{\BIBentryALTinterwordstretchfactor}{4}
\providecommand{\BIBentryALTinterwordspacing}{\spaceskip=\fontdimen2\font plus
\BIBentryALTinterwordstretchfactor\fontdimen3\font minus
  \fontdimen4\font\relax}
\providecommand{\BIBforeignlanguage}[2]{{%
\expandafter\ifx\csname l@#1\endcsname\relax
\typeout{** WARNING: IEEEtran.bst: No hyphenation pattern has been}%
\typeout{** loaded for the language `#1'. Using the pattern for}%
\typeout{** the default language instead.}%
\else
\language=\csname l@#1\endcsname
\fi
#2}}
\providecommand{\BIBdecl}{\relax}
\BIBdecl

\bibitem{zhou2020near}
F.~Zhou and Y.~Chai, ``Near-sensor and in-sensor computing,'' \emph{Nature
  Electronics}, vol.~3, no.~11, pp. 664--671, 2020.

\bibitem{chen2018scamp5d}
J.~Chen, S.~J. Carey, and P.~Dudek, ``Scamp5d vision system and development
  framework,'' in \emph{Proceedings of the 12th International Conference on
  Distributed Smart Cameras}, 2018, pp. 1--2.

\bibitem{carey2013100}
S.~J. Carey, A.~Lopich, D.~R. Barr, B.~Wang, and P.~Dudek, ``A 100,000 fps
  vision sensor with embedded 535gops/w 256$\times$ 256 simd processor array,''
  in \emph{2013 Symposium on VLSI Circuits}.\hskip 1em plus 0.5em minus
  0.4em\relax IEEE, 2013, pp. C182--C183.

\bibitem{simons2019review}
T.~Simons and D.-J. Lee, ``A review of binarized neural networks,''
  \emph{Electronics}, vol.~8, no.~6, p. 661, 2019.

\bibitem{liu2020bmvc}
Y.~Liu, L.~Bose, J.~Chen, S.~J. Carey, P.~Dudek, and W.~Mayol-Cuevas,
  ``High-speed light-weight cnn inference via strided convolutions on a pixel
  processor array,'' in \emph{The 31st British Machine Vision Conference
  (BMVC)}, 2020.

\bibitem{bose2020}
L.~Bose, P.~Dudek, J.~Chen, S.~J. Carey, and W.~W. Mayol-Cuevas, ``Fully
  embedding fast convolutional networks on pixel processor arrays,'' in
  \emph{European Conference on Computer Vision -- ECCV 2020}.

\end{thebibliography}
\end{document}